\documentclass{article}

\usepackage[preprint]{neurips_2019}

\usepackage[utf8]{inputenc} 
\usepackage[T1]{fontenc}    
\usepackage{hyperref}       
\usepackage{url}            
\usepackage{booktabs}       
\usepackage{amsfonts}       
\usepackage{nicefrac}       
\usepackage{microtype}      

\usepackage{amsmath,amssymb}
\usepackage{graphicx}
\usepackage{subfigure}
\usepackage{caption}

\usepackage{multirow}
\usepackage{makecell}

\newcommand{\bb}[1]{\mathbf{#1}}

\newcommand{\bx}{\bb{x}}
\newcommand{\by}{\bb{y}}
\newcommand{\bh}{\bb{h}}
\newcommand{\bs}{\bb{s}}
\newcommand{\bw}{\bb{w}}
\newcommand{\bz}{\bb{z}}
\newcommand{\bmu}{\boldsymbol{\mu}}

\title{Generative Model with Dynamic Linear Flow}

\author{%
  Huadong Liao$^\dagger$\And Jiawei He$^\ddagger$ \And Kunxian Shu$^\dagger$\\
  \AND\\
  $^\dagger$Chongqing University of Posts and Telecommunications, Chongqing, China\\
  $^\ddagger$School of Life Sciences, Jilin University, China\\
  $^\ddagger$Zhuhai College of Jilin University, Zhuhai, China
}

\begin{document}

\maketitle

\begin{abstract}
  Flow-based generative models are a family of exact log-likelihood models with tractable
  sampling and latent-variable inference, hence conceptually attractive for modeling
  complex distributions. However, flow-based models are limited by density estimation performance
  issues as compared to state-of-the-art autoregressive models. Autoregressive models, which
  also belong to the family of likelihood-based methods, however suffer from
  limited parallelizability. In this paper,
  we propose \emph{Dynamic Linear Flow (DLF)}, a new family of invertible transformations
  with partially autoregressive structure. Our method benefits from the efficient computation of flow-based methods
  and high density estimation performance of autoregressive methods. We demonstrate that the proposed
  DLF yields state-of-the-art performance on ImageNet 32$\times$32 and 64$\times$64 out of all flow-based methods,
  and is competitive with the best autoregressive model.
  Additionally, our model converges 10 times faster than Glow~\citep{kingma2018glow}. The code is available at \url{https://github.com/naturomics/DLF}.
\end{abstract}

\section{Introduction}
\label{sec:introduction}

The increasing amount of data, paired with the exponential progress
in the capabilities of hardware and relentless efforts for better methods,
has tremendously advanced the development in the fields of deep learning, such as
image classification~\citep{krizhevsky2012imagenet,he2016deep,huang2017densely}
and machine translation~\citep{vaswani2017attention,devlin2018bert,radford2019language}.
However, most applications have been greatly limited to situations
where large amounts of supervision is available, as labeling data remains a labor-intensive
and cost-inefficient exercise. In the meantime, unlabeled data is generally easier to acquire but its
direct utilization is yet a central challenging problem.
Deep generative models, an emerging and popular branch of machine learning, aims to address these
challenges by modeling the high-dimensional distributions of data without supervision.

In recent years, the field of generative modeling has advanced significantly, especially in the development and application of generative adversarial networks (GANs)~\citep{goodfellow2014generative}
and likelihood-based methods~\citep{graves2013generating,kingma2013auto,dinh2014nice,oord2016pixel}.
Likelihood-based generative methods could be further divided
into three different categories: variational autoencoders~\citep{kingma2013auto},
autoregressive models~\citep{oord2016pixel,salimans2017pixelcnn++,chen2017pixelsnail,menick2018generating},
and flow-based generative methods~\citep{dinh2014nice,dinh2016density,kingma2018glow}.
Variational autoencoders have displayed promising parallelizability of training
and synthesis, however, it could be technically challenging to optimize with the lower
bound on the marginal likelihood of the data. Autoregressive models and
flow-based generative models both estimate the exact likelihood of the data.
However, autoregressive models suffer from the limited parallelizability of
synthesis or training, and a lot of effort has been made to overcome this
drawback~\citep{oord2017parallel}.
On the contrast, flow-based generative models are efficient for training and
synthesis, but generally yield compromised performance in comparison with autoregressive models in density
estimation benchmarks.

In this paper, we focus on the exact likelihood-based methods.
In Section~\ref{background}, we first review models of autoregressive
methods and flow-based methods. Inspired by their common properties, in Section~\ref{method},
we then propose a new family of invertible transformations with partially
autoregressive structure. And we illustrate that autoregressive models and flow-based
generative models are two extreme forms of our proposed method.
In Section~\ref{experiments}, our empirical results show that the proposed method achieves
state-of-the-art density estimation performance on ImageNet dataset among flow-based methods and converges significantly
faster than Glow model~\citep{kingma2018glow}. Though our method has a partially autoregressive structure,
we illustrate that the synthesis of a high-resolution image (i.e., 256$\times$256 image) on modern hardware takes less than one second, which is comparable to most flow-based methods.

\section{Background}
\label{background}

\subsection{Flow-based Models}
In most flow-based models~\citep{kingma2018glow,dinh2014nice,dinh2016density},
the high-dimensional random variable $\bx$ with complex
and unknown true distribution $\bx\sim p^\star(\bx)$ is generally modeled by a latent variable
$\bz$: $\bz = f_\theta(\bx)$, where $f$ can be any bijective function with parameters $\theta$ and
is typically composed of a series of transformations
$f=f_1\circ f_2\circ\cdots\circ f_L$. $p_{\theta}(\bz)$ has a tractable
density, such as a standard Gaussian distribution.
With the change of variables formula, we then have the marginal log-likelihood of a datapoint
and take it as the optimization objective of learning $\theta$:
\begin{equation}\begin{aligned}
\log p_{\theta}(\bx)
&= \log p_{\theta}(\bz) + \log|\det({d\bz}/{d\bx})| \\
&=\log p_{\theta}(\bz) + \sum_{i=1}^L\log|\det({d\bh_i}/{d\bh_{i-1}})|
\end{aligned}\end{equation}
where $\bh_i = f_i(\bh_{i-1})$ is the hidden output of sequence of transformations, with
$\bh_0 \triangleq \bx$ and $\bh_L \triangleq \bz$.

However, the above formula requires
the computation of Jacobian determinant of each intermediate transformation, which is
generally intractable and therefore, becomes a limitation of the above method.
In practice, to overcome this issue, the transformation
function $f_i$ is well-designed to let its Jacobian matrix be triangular or diagonal, thus
the log-determinant is simply the sum of log-diagonal entries:
\begin{equation}
\log|\det(d\bh_i/d\bh_{i-1})| =
\text{sum}(\log|\text{diag}(d\bh_i/d\bh_{i-1})|)
\end{equation}

In the next part of this section, we will review invertible and tractable
transformations reported in previous studies, categorized
as fully autoregressive structure and non-autoregressive structure. After that, we will discuss their respective  advantages  and  disadvantages
in computational parallelizability and density estimation performance.

\subsection{Autoregressive and Inverse Autoregressive Transformations}
\cite{papamakarios2017masked} and \cite{kingma2016improved} introduced autoregressive (AR)
transformation and Inverse Autoregressive (IAR) transformation, respectively. These
methods model a similar invertible and tractable transformation from high-dimensional variable $\bx$ to $\by$:
\begin{equation}
y_i = s_ix_i + \mu_i
\end{equation}
where $x_i$ and $y_i$ are the $i$-th element of $\bx$ and $\by$, respectively.
The difference between AR and IAR is that $s_i$ and $\mu_i$ are driven by different
input: $s_i, \mu_i = g(\bx_{1:i-1})$ in autoregressive transformation and
$s_i, \mu_i = g({\by}_{1:i-1})$ in inverse autoregressive transformation. Here $g$
is an arbitrarily complex function, usually a neural network.
The vectorized transformation and its reverse transformation for (inverse) autoregressive
transformations could be described as follows:
\begin{align}
f:\ &\by=\bs\odot\bx+\bmu\\
f^{-1}:\ &\bx=(\by-\bmu)/\bs \label{reversedf}
\end{align}
where $\odot$ is the Hadamard product or element-wise product, and the addition,
division and subtraction are also element-wise operations.

In previous works, AR and IAR have been successfully applied to image generation~\citep{kingma2016improved}
and speech synthesis~\citep{oord2016wavenet}. However, as $s_i, \mu_i$ are dependent on previous
elements of input $\bx_{1:i-1}$ or output $\by_{1:i-1}$, these transformations are inherently
sequential in at least one pass of training (IAR) or synthesis (AR),
making it difficult to parallelize on modern parallel hardware~\citep{oord2017parallel}.

\subsection{Non-autoregressive Transformations}
Non-autoregressive transformations are designed to be parallelizable in both forward and backward pass, with tractable
Jacobian determinants and inverses. Here, we describe a number of them:

\textbf{Actnorm}~\citep{kingma2018glow}, 
as one of non-autoregressive transformations, was proposed to
alleviate the training problems encountered in deep models, which is actually a
special case of (inverse) autoregressive transformation that the scale
$\bs$ and bias $\bmu$ are treated as regular trainable parameters, namely,
independent of the input data:
\begin{equation}
{\by}={\bs}\odot{\bx}+\bmu\text{ where }{\bs}\text{ and }\bmu \text{ are learnable}
\end{equation}
It's worth mentioning that $\bs$ and $\bmu$ are shared between the spatial dimensions of
$\bx$ when the input is 2D images as described in~\cite{kingma2018glow}.

\textbf{Affine/additive coupling layers}~\citep{dinh2014nice,dinh2016density} split the high-dimensional input
$\bx$ into two parts $(\bx_1, \bx_2)$ and applies different transformations to each one
to obtain the output $\by=(\by_1, \by_2)$.
The first part is transformed with an identity function thus remains unchanged,
and the second part is mapped to a new distribution with an affine transformation:
\begin{equation}\begin{aligned}
\by_1&=\bx_1 \\
\by_2&=\bx_2\odot\bs+\bmu \label{affLayer}
\end{aligned}\end{equation}
with $\bmu, \bs=g(\bx_1)=g(\by_1)$. Same as AR and IAR, here $g$ is an arbitrarily
complex function, typically a neural network.
Note that this transformation can be also rewritten in the same form as
(inverse) autoregressive transformations and actnorm method:
$\by=\bs^\prime\odot\bx+\bmu^\prime$,
where $\bs^\prime=[1, \bs]^T$ and $\bmu^\prime=[0, \bmu]^T$.

These non-autoregressive transformations have the advantage of parallelization,
therefore, they are usually faster than the transformations with autoregressive structure.
However, previous results have shown that they generally perform much worse in density
estimation benchmarks~\citep{ho2019flow++}.

\section{Method}
\label{method}

In this section, we introduce a new family transformations, which have the advantages of
computational efficiency of non-autoregressive transformations and the high performance
of (inverse) autoregressive transformations in density estimation benchmarks.

There are two key observations from the mentioned methods in Section~\ref{background}. First, all methods have a consistent linear form:
\begin{equation}
\by=\bw\bx+\bmu,\text{ where }\bw=\text{diag}(\bs)
\end{equation}
Here $\bw$ is a diagonal matrix with $\bs$ as its diagonal elements, thus this
transformation is invertible and its inverse is simple as Eq.~\eqref{reversedf}.
The invertibility makes it possible to use a same transformation as the block of both
encoder and decoder in generative models.

The second key observation is the weights of such linear transformations $\bw$ and $\bmu$
are data-dependent, in the way that the determinant of Jacobian matrix
$J=d\by/d\bx$ is computationally efficient or tractable, usually making $J$
triangular (AR, IAR and affine coupling layer) or diagnoal (actnorm).
Therefore, the log-determinant is simply the sum of logarithm of diagonal terms
$\log(\det(J)) = \text{sum}(\log|\bs|)$.
Their difference are the methods used for modelling the relationship between the
weights $(\bw,\bmu)$ and the data under the "easy determinant of the Jacobian" constraint.

\begin{figure}[t]
  \centering
  \includegraphics[width=0.9\textwidth]{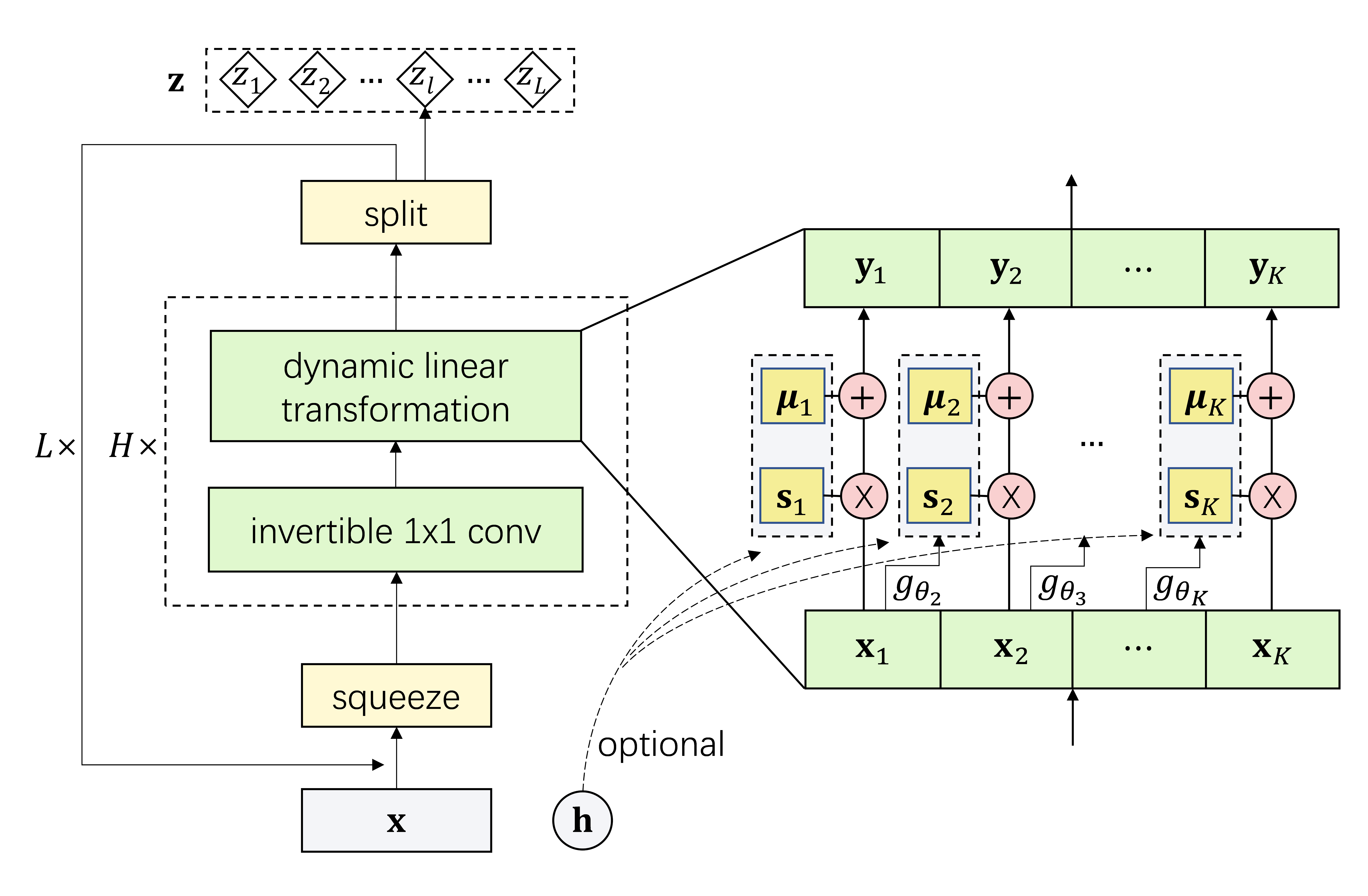}
  \caption{Dynamic Linear Flow with multi-scale architecture. At each scale, the input is passed through a squeezing
  operation to trade the spatial size for number of channels, followed by $H$ flows of invertible 1$\times$1 convolution
  and dynamic linear transformation. The output is splitted into two halves, one for the next series of flow and another
  as a part of final latent variable. The condition $\bh$ is optional which guides dynamic linear transformation as prior knowledge.}
  \label{fig:dlf}
\end{figure}

\subsection{Dynamic Linear Transformation with Triangular Jacobian}
\label{subsec:dlf}

Let us now consider a high-dimensional variable $\bx\in R^D$: When splitting it into $K$ parts
along its dimension, we obtain $\bx=(\bx_1,\dots,\bx_K)$, with $1\le K \le D$.
Then we introduce a tractable and bijective function $\by=f(\bx)$ as following:
\begin{equation}\begin{aligned}
\by_1 &= h(\bx_1) \\
\by_k &= \bs(\bx_{k-1})\odot\bx_k+\bmu(\bx_{k-1})
\end{aligned} \label{Eq:dynamiclinear}\end{equation}
with $k=2,\cdots,K$. Variables $\by_k,\bx_k,\bs_k$ and $\bmu_k$ have the same dimension, and
$\bs_k,\bmu_k=g_{\theta_{k}}(\bx_{k-1})$ are modeled by an arbitrarily complex function (usually a neural network) with
the previous part of data as input.
$h()$ is tractable and bijective with the inverse $\bx_1=h^{-1}(\by_1)$. An alternative of $h()$ is identity
function $\by_1=\bx_1$. If then, combined with $K=2$, our method turns out to be
the case of affine coupling layer, see Eq.~\eqref{affLayer}. For the purpose of consistency, in this paper,
we choose $h(\bx_1)=\bs_1\odot\bx_1+\bmu_1$, where $\bs_1$ and $\bmu_1$ are trainable.
In other words, $\bs_1$ and $\bmu_1$ are modeled by $g(\bx_0)$ with that $\bx_0$
is any constant, e.g.\ $\bx_0=1$.
Therefore, Eq.~\eqref{Eq:dynamiclinear} and its inverse can be rewritten as:
\begin{align}
f:\ \by_k &= \bs(\bx_{k-1})\odot\bx_k+\bmu(\bx_{k-1})\\ 
f^{-1}:\ \bx_k &= (\by_k-\bmu(\bx_{k-1}))/\bs(\bx_{k-1})
\end{align}
where $k=1,2,\cdots,K$ and initial condition $\bx_0=1$.

The Jacobian of the above transformation is triangular with $\bs=(\bs_1,\cdots,\bs_K)$ as its diagonal
elements and thus has a simple log-determinant term:
\begin{equation}
  \log(|\det(d\by/d\bx)|) = \sum_{k=1}^{K}\text{sum}(\log|\bs_k|)\label{logdet}
\end{equation}

Note that our proposed transformation can also be rewritten in the following linear form:
\begin{equation}\begin{bmatrix}
1 \\ \by
\end{bmatrix} = \begin{bmatrix}
1&0 \\ \bb{b} & \bw
\end{bmatrix}\begin{bmatrix}
1 \\ \bx
\end{bmatrix}\end{equation}
where the variables $\bb{b}^T=\big(\bmu(\bx_0)^T, \cdots,\bmu(\bx_{K-1})^T\big)$ and
$\bw= \text{diag}\big([\bs(\bx_0)^T,\cdots, \bs(\bx_{K-1})^T]\big)$, and they are data-dependent, therefore,
we call our method \emph{dynamic linear transformation}. As $\bw$ and $\bb{b}$ changed for different inputs, dynamic linear
transformation can be considered as the extreme form of piecewise linear function, each of the
points learning its own weights for affine transformation.

\begin{figure}[t]
  \centering
  \includegraphics[width=0.8\textwidth]{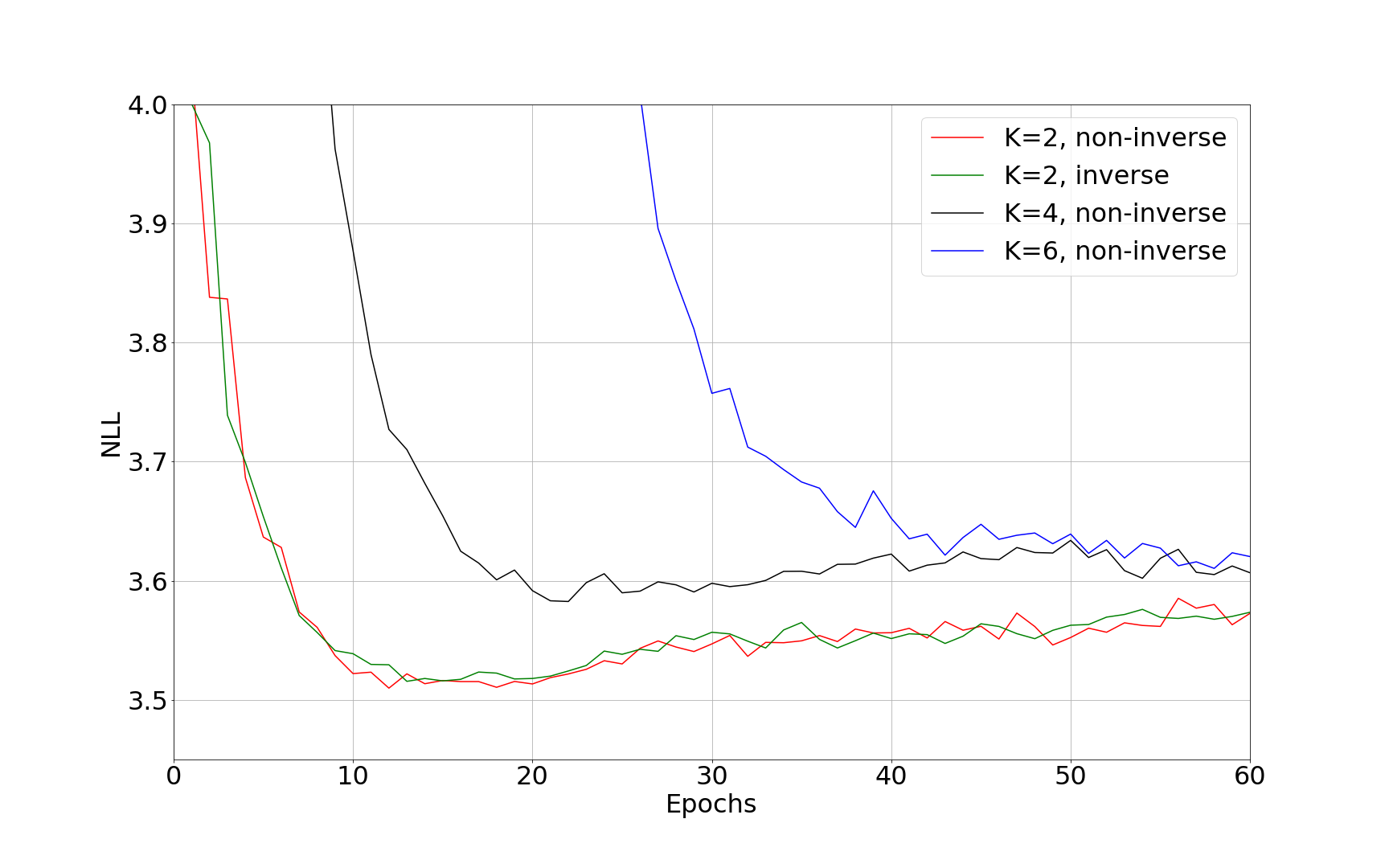}
  \caption{Negative log-likelihood on CIFAR-10 test set during training. Increasing $K$ leads to no performance gain but slower convergence.}
  \label{fig:loss}
\end{figure}

In applications, an important concern for dynamic linear transformation is its recursive
dependencies in the reverse pass, introduced by that each pair $(\bs_k,\bmu_k)$ depends
on previous partition $\bx_{k-1}$. We show that this issuse could be addressed for two
reasons: (1) the recursive dependencies are based on piece and only dependent of one earlier step,
thus it is more efficient on computation than the element-level autoregressive structure,
which has a great dependency on all its earlier steps;
and (2) the smaller $K$ is, the shorter the dependency chain we get.
In Section~\ref{experiments}, we will show that increasing $K$ is not helpful and results in worse
NLL score (Fig.~\ref{fig:loss}), and our state-of-the-art results are achieved
with $K=2$, with a similar computational speed compared to non-autoregressive methods.

Similar to the transformations of AR and IAR, we also introduce a variant of dynamic linear transformation.
Let $\bs_k()$ and $\bmu_k()$ take the transformed output $\by_{k-1}$ as input
instead of $\bx_{k-1}$, we then have:
\begin{align}
f:\ \by_k &= \bs(\by_{k-1})\odot\bx_k+\bmu(\by_{k-1}) \\
f^{-1}:\ \bx_k &= (\bx_k-\bmu(\by_{k-1}))/\bs(\by_{k-1})
\end{align}
with $k=1,2,\cdots,K$ and initial condition $\by_0=1$. We call this variant
\emph{inverse dynamic linear transformation}, which has the same
log-determinant as Eq.~\eqref{logdet}.

\subsection{Conditional Dynamic Linear Transformation}

In most samples generation scenarios, it is a common requirement to control
the generating process with prior knowledge, e.g.\ generating an image with class
label information. We introduce the conditional dynamic linear transformation
to meet such requirement. Given condition $\bh$, the conditional dynamic linear
transformation could be described as:
\begin{align}
\by_k &= \bs(\bx_{k-1},\bh)\odot\bx_k + \bmu(\bx_{k-1},\bh)
\end{align}
The parameters of transformation $\bs_k$ and $\bmu_k$ take $\bh$ as an additional input.
Accordingly, when inverting the transformation, we can recompute $\bs_k$ and $\bmu_k$
from the same $\bh$ and transformed $\bx_{k-1}$.

For the inverse dynamic linear transformation variant, its conditional form is
\begin{align}
\by_k &= \bs(\by_{k-1},\bh)\odot\bx_k + \bmu(\by_{k-1},\bh)
\end{align}

\section{Dynamic Linear Flow}
\label{DLF}

In high-dimensional problems (e.g. generating images of faces), the use of a single layer of
dynamic linear transformation is fairly limited. In order to increase the capability of the
model, in this section, we describe Dynamic Linear Flow (DLF), a flow-based model using
the (inverse) dynamic linear transformation as a building block. Following by the previous
works of NICE~\citep{dinh2014nice}, RealNVP~\citep{dinh2016density} and
Glow, DLF is stacked with blocks consisting of
invertible $1 \times 1$ convolution and (inverse) dynamic linear transformation,
combined in
a multi-scale architecture (Fig.~\ref{fig:dlf}).
Since dynamic linear transformation and inverse dynamic linear
transformation are similar, in Fig.~\ref{fig:dlf}, we only illustrate the structure of DLF
with dynamic linear transformation, and the corresponding variant is obtained by
replacing the layer of dynamic linear transformation with inverse dynamic linear transformation. A comparison on their density estimation performance is included in Section~\ref{experiments}.

\subsection{Multi-scale Architecture}
For the case of 2D image input, following realNVP and Glow, we use
squeezing operation to reduce each spatial resolution by a factor 2 and transpose them into
channels, resulting in $s \times s \times c$ input transformed into a
$\frac{s}{2} \times \frac{s}{2} \times 4c$ tensor. After the squeezing operation, $H$ steps of
flows consisting of invertible $1 \times 1$ convolution and dynamic linear
transformation are combined into a sequence. Then the output of sequence stacks is factored
out half of the dimensions at regular intervals, while all of the another half at different
scales are concatenated to obtain the final
transformed output. The above operations are iteratively applied for $L$ times.

\subsection{Invertible $1 \times 1$ Convolution}
To ensure that each dimension can influence every other dimension during the transformation,
we apply an invertible $1 \times 1$ convolution layer~\citep{kingma2018glow}
before each layer of dynamic linear transformation. The invertible $1 \times 1$ convolution
is essentially a normal $1 \times 1$ convolution with equal number of input and output channels:
\begin{equation}
  \forall i,j: \by_{i,j} = \mathbf{W} \bx_{i,j} \\ 
\end{equation}
where $\mathbf{W}$ is the kernel with shape $c \times c$, and $i,j$ index the spatial dimension of 2D variables $\bx,\by$.

\begin{table}[t]
  \centering
  \caption{Hyperparameters and number of trainable parameters for our experiments in Section~\ref{experiments}.}
  \begin{tabular}{ccccc}
    \toprule
    \textbf{Dataset} & \multicolumn{1}{c}{\bf \makecell{Partitions ($K$)}} & \multicolumn{1}{c}{\bf \makecell{Channels ($c$)}} & \multicolumn{1}{c}{\bf\makecell{Levels ($L$)}} & \multicolumn{1}{c}{\bf\makecell{parameters}} \\
    \midrule
    \multirow{3}{*}{CIFAR-10} & 2 & 512 & 3 & 44.6M \\
                   & 4 & 308 & 3 & 45.5M \\
                      & 6 & 246 & 3 & 45.7M \\
    MNIST & 2 & 128 & 2 & 1.8M\\
    ImageNet 32$\times$32     & 2 & 512 & 3 & 44.6M \\
    ImageNet 64$\times$64     & 2 & 384 & 4 & 50.7M \\
    CelebA HQ 256$\times$256  & 2 & 128 & 6 & 57.4M \\
    \bottomrule
  \end{tabular}
  \label{hyperparams}
\end{table}

\section{Experiments}
\label{experiments}

We evaluate the proposed DLF model on standard image modeling benchmarks such as
CIFAR-10~\citep{krizhevsky2009learning}, ImageNet~\citep{russakovsky2015imagenet} among others.
We first investigated the impact of number of partitions $K$ and compared the variants of
dynamic linear transformation. With the optimal hyperparameters, we then compared
log-likelihood with previous generative models of autoregressive and
non-autoregressive families. Lastly, we assessed the conditional DLF with class label information
and the qualitative aspects of DLF on high-resolution datasets.

In all our experiments, we followed a similar implementation of neural network $g_{\theta_k}$
as in Glow, using three convolutional layers with a different activation function in the
last layer. More specifically, the first two convolutional layers have $c$ channels
with ReLU activation functions, and $3 \times 3$ and $1 \times 1$ filters, respectively. To control the number of model parameters, $c$ varied for different number of partitions
$K$ and different datasets (Table.~\ref{hyperparams}). The last convolution is $3 \times 3$
and has two times of channels as partition $\bx_k$, and its outputs $o$ are equally
splitted into two parts along the channel dimension, obtained $\log\bs_k^\prime,\bmu_k=\text{split}(o)$.
For the purpose of training stability, the final $\bs_k=\exp(\alpha\tanh(\log\bs_k^\prime)+\beta)$, where
$\alpha$ and $\beta$ are learnable scale variables. For the conditional DLF,
we introduce conditions by $\log\bs_k^\prime, \bmu_k=\text{split}(o+Vh)$ in the last layer,
where $V$ is weight matrix for conditioning data. In cases
where $h$ encodes spatial information, the matrix products ($Vh$) is replaced by a
$3 \times 3$ convolution operation. The parameters $\theta_k$ of neural network are individual
between different partitions $\bx_k$. Depth $H$ is always set to 32.
See Table.~\ref{hyperparams} and Appendix~\ref{appendix:optim} for more details of optimization.

\begin{table}[t]
  \caption{Comparison on density estimation performance (bits/dim, lower is better). Results are obtained from 8-bits datasets.}
  \label{results}
  \centering
  \resizebox{\textwidth}{!} {
  \begin{tabular}{clccc}
    \toprule
    \textbf{Family} & \textbf{Model} & \textbf{CIFAR10} & \multicolumn{1}{c}{\bf \makecell{ImageNet \\32$\times$32}} & \multicolumn{1}{c}{\bf\makecell{ImageNet \\64$\times$64}} \\
    \midrule
    \multirow{4}{*}{Non-autoregressive}
        & RealNVP~\citep{dinh2016density} & $3.49$ & $4.28$ & $3.98$ \\
        & Glow~\citep{kingma2018glow} & $3.35$ & $4.09$ & $3.81$ \\
        & Flow++~\citep{ho2019flow++} & $3.09$ & $3.86$ & $3.69$ \\
        & \textbf{DLF (ours)}  & $\mathbf{3.44}$ & $\mathbf{3.85}$ & $\mathbf{3.57}$ \\
    \midrule
    \multirow{5}{*}{Autoregressive}
        & Multiscale PixelCNN~\citep{reed2017parallel} & - & 3.95 & 3.70 \\
        & PixelRNN~\citep{oord2016pixel} & 3.00 & 3.86 & 3.63 \\
        & Gated PixelCNN~\citep{van2016conditional} & 3.03 & 3.83 & 3.57 \\
        & PixelSNAIL~\citep{chen2017pixelsnail} & 2.85 & 3.80 & 3.52 \\
        & SPN~\citep{menick2018generating} & - & 3.79 & 3.52 \\
    \bottomrule
  \end{tabular}}
  \label{densityEstimation}
\end{table}

\subsection{Effect of Partitions $K$ and Model Variants}
Choosing a large $K$ will increase the recursive complexity of the model. Therefore, a small
$K$ is preferred given the performance was not degraded. We
tested number of partitions $K=2,4$ and $6$ on CIFAR-10. The number of model parameters
was approximately equal to 45M (same size as in Glow) by controlling channels $c$,
see Table~\ref{hyperparams}.
The results are summarized in Fig.~\ref{fig:loss}. As we can see, Increasing $K$ is unnecessary and has negative effect
on model performance, leading to worse NLL score and slower convergence.
On the other hand, we replaced the layers of dynamic linear transformation with its inverse
variant when $K=2$, which does not produce significant performance difference.
Therefore, we choose $K=2$ and will not evaluate DLF with inverse dynamic linear
transformation in the following experiments.

Note that for the case of $K=2$, both the non-inverse and inverse variants start overfitting
after 20 epochs. And after 50 epochs, the averaged NLL score over epoch on training set reaches 3.30 and
the loss still keeps decreasing, while the validation NLL increases from 3.51 to 3.55.
As mentioned in Section~\ref{method}, dynamic linear transformation is the extreme
form of piecewise linear function, learning weights of affine transformation for each input.
This indicates that the more powerful the transformation is, the more training data our method is
eager for to cover the distribution of whole dataset.
Therefore, to avoid overfitting,
apart from degrading the capacity of dynamic linear transformation, another approach is to increase the size of
training dataset. We will discuss this in greater details in the following sections.

\begin{figure}[t]
  \centering
  \subfigure[Unconditional samples]{\includegraphics[width=0.45\linewidth]{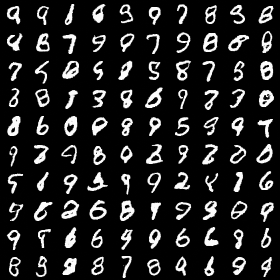}}
  \hspace{2.5mm}
  \subfigure[Conditional samples]{\includegraphics[width=0.45\linewidth]{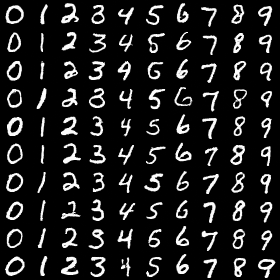}}
  \caption{Comparison of unconditional and conditional DLF on MNIST with class label information. (a) unconditional samples; (b) class conditional samples. Temperature 0.7}
  \label{fig:mnist}
\end{figure}

\subsection{Density Estimation}
To compare with previous likelihood-based models, we perform density estimation on
natural images datasets CIFAR10 and ImageNet.
In particular, we use the $32\times 32$ and $64\times 64$ downsampled version of
ImageNet~\citep{oord2016pixel}. For all datasets, we follow the same preprocessing
as in~\cite{kingma2018glow}. 

On CIFAR10, as discussed earlier, the DLF model with the same size as Glow displayed overfitting. A possible reason is the simplicity and small size
of CIFAR10. We tested the assumption by training a same size model on the relatively complex dataset
ImageNet 32$\times$32. As shown in Table.~\ref{densityEstimation}, compared to Glow, the improvement
is significant by 0.24 bits/dim and we did not observe overfitting on
Imagenet 32$\times$32. This encourages us to apply transfer learning to CIFAR10, initializing its
parameters with the trained model on ImageNet 32$\times$32. We found the approach helpful for CIFAR10,
obtained 3.51 bits/dim without transfer learning and 3.44 bits/dim with transfer learning.
on ImageNet 64$\times$64, the DLF model led to 3.57 bits/dim, while the model size is
relatively small with 50.7M parameters compared to 112.3M parameters of Glow on the same dataset.

Summarily, the DLF model achieves state-of-the-art density modeling results on ImageNet 32$\times$32 and 64$\times$64 among all non-autoregressive models, and it is comparable to most autoregressive models. It is worth mentioning that all results are obtained within 50 epochs. To our knowledge, it is more than 10 times more efficient than Glow and Flow++~\citep{ho2019flow++}, which generally require at least thousands of epochs to converge.

\subsection{Conditional DLF}
For conditional DLF, we experimented on MNIST~\citep{lecun1998gradient} and CIFAR10 with
class label as prior. The hyperparameters can be found in Table.~\ref{hyperparams}
(For CIFAR10, only $K=2$ was tested). For the conditional version, during
training, we represent the class label as a 10-dimensional, one-hot encoded vector $\bh$,
and add it to each layer of dynamic linear transformation. On contrary, class label is not given in the unconditional
version. Once converged,
we synthesize samples by randomly generating latent variables $\bz$ from standard Gaussian
distribution, and giving one-hot encoded label to all layers of dynamic linear
transformation for conditional DLF. As in Fig.~\ref{fig:mnist},
the class-conditional samples (sampled after 150 epochs)
are controlled by the corresponding label and the quality is better than the unconditional samples
(sampled after 200 epochs). This result indicates that DLF correctly learns to control the distribution
with class label prior. See appendix for samples from CIFAR10.

\begin{figure}[t]
  \centering
  \includegraphics[width=0.4\textwidth]{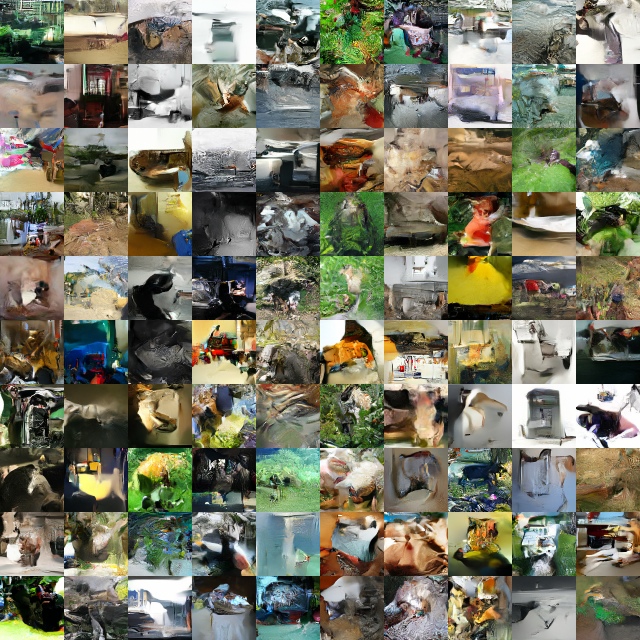}
  \hspace{2.5mm}
  \includegraphics[width=0.53\textwidth]{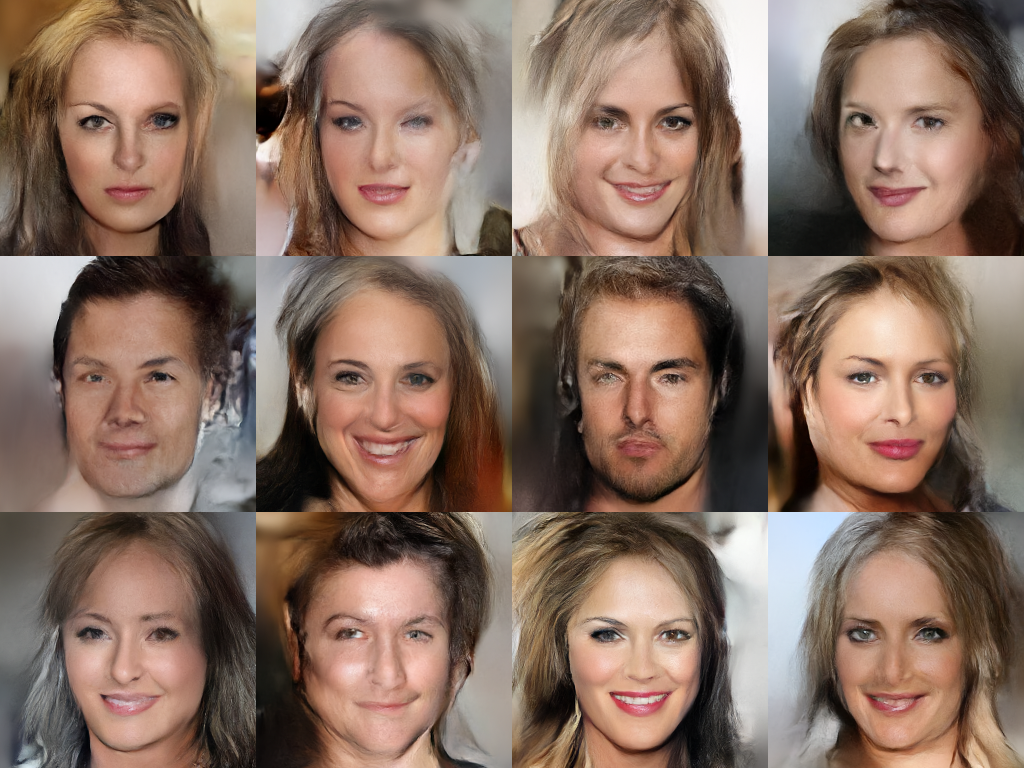}
  \caption{Random samples from ImageNet 64$\times$64 (left, temperature 1.0) and CelebA-HQ 256$\times$256 (right, temperature 0.6), both on 8-bits.}
  \label{fig:imgnet64}
\end{figure}

\subsection{Samples and  Interpolation}
We present samples randomly generated from the trained DLF model on ImageNet 64$\times$64 and CelebA HQ
256$\times$256~\citep{karras2017progressive} in Fig.~\ref{fig:imgnet64}, both on 8-bit.
For CelebA 256$\times$256 dataset, our model has 57.4M parameters, which is approximately $1/4$ of Glow's,
and is trained with only 400 epochs. Note that our model have not fully converged on CelebA 256$\times$256,
due to limited computational resources.

\begin{figure}[t]
  \centering
  \includegraphics[width=0.9\textwidth]{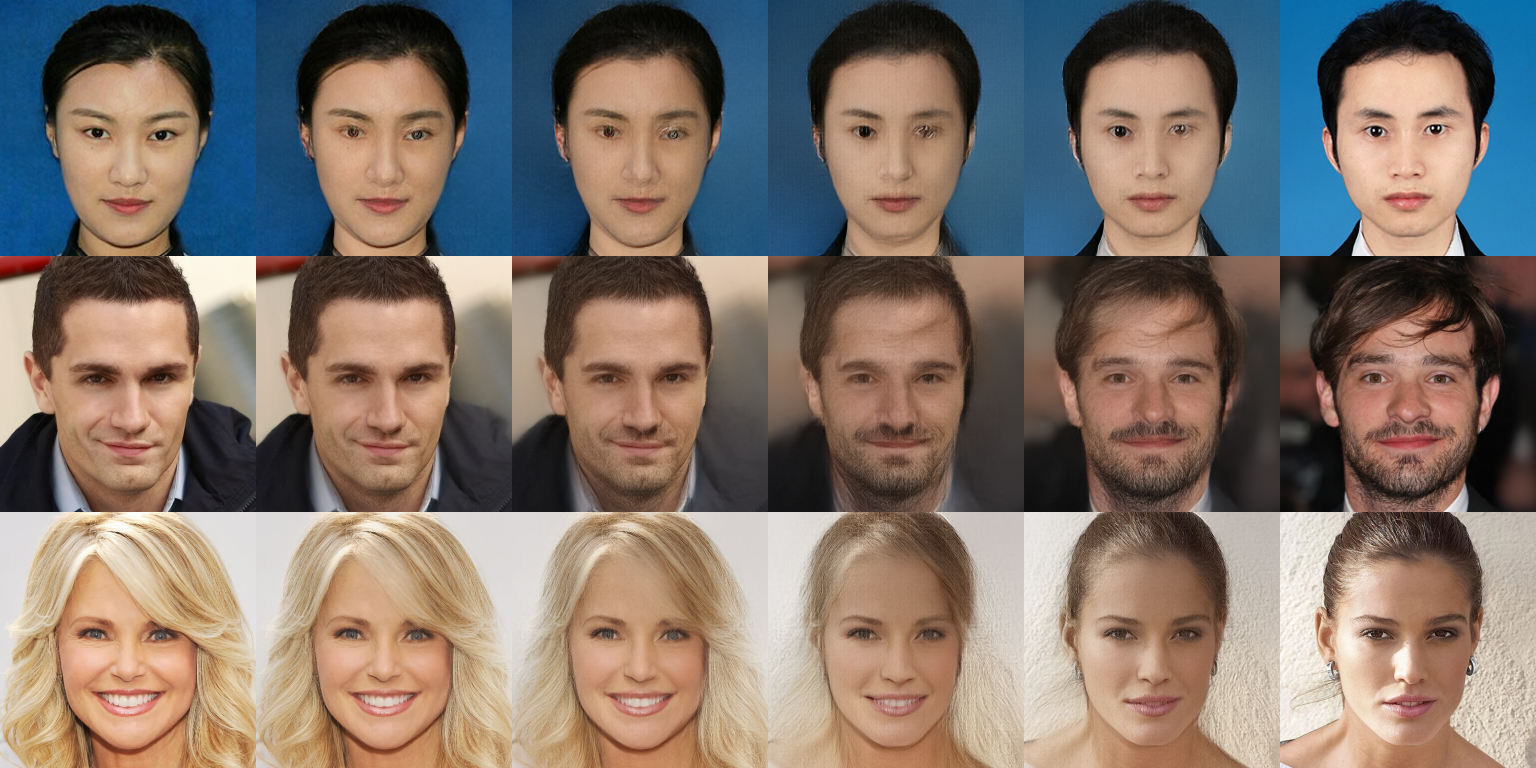}
  \caption{Linear interpolation in latent space between two real images.
  The images have never been seen by model during training.}
  \label{fig:interpolation}
\end{figure}

In Fig.~\ref{fig:interpolation}, we take pairs of real images from Celeba HQ 256$\times$256 test set, encode them to obtain the latent
representations, and linearly interpolate between the latents to decode samples. As we can see, the image manifold is smoothly changed.

During sampling, generating a 256$\times$256 image at batch size 1 takes about 315ms on a single 1080 Ti GPU,
and 1078ms on a single i7-6700k CPU. We believe this sampling speed can be further improved by using inverse
dynamic linear transformation, as it has no recursive structure in the reverse computation.

\section{Conclusion}
We propose a new family of invertible and tractable transformations, coined \emph{dynamice
linear transformation}. Building DLF model with blocks of dynamic linear transformation,
we achieved state-of-the-art performance in terms of log-likelihood on ImageNet
32$\times$32 and 64$\times$64 benchmarks. We also illustrated that our flow-based model can
efficiently synthesize high-resolution images.

Flow-based methods optimize exact log-likelihood directly, which is stable and easy for
training. With the development of more powerful invertible transformations, we belief
flow-based methods will show potential comparable to GANs and give rise to various applications.

%

\medskip

\small
\bibliographystyle{apalike}
\bibliography{bib}

\begin{thebibliography}{}

\bibitem[Chen et~al., 2017]{chen2017pixelsnail}
Chen, X., Mishra, N., Rohaninejad, M., and Abbeel, P. (2017).
\newblock Pixelsnail: An improved autoregressive generative model.
\newblock {\em arXiv preprint arXiv:1712.09763}.

\bibitem[Devlin et~al., 2018]{devlin2018bert}
Devlin, J., Chang, M.-W., Lee, K., and Toutanova, K. (2018).
\newblock Bert: Pre-training of deep bidirectional transformers for language
  understanding.
\newblock {\em arXiv preprint arXiv:1810.04805}.

\bibitem[Dinh et~al., 2014]{dinh2014nice}
Dinh, L., Krueger, D., and Bengio, Y. (2014).
\newblock Nice: Non-linear independent components estimation.
\newblock {\em arXiv preprint arXiv:1410.8516}.

\bibitem[Dinh et~al., 2016]{dinh2016density}
Dinh, L., Sohl-Dickstein, J., and Bengio, S. (2016).
\newblock Density estimation using real nvp.
\newblock {\em arXiv preprint arXiv:1605.08803}.

\bibitem[Goodfellow et~al., 2014]{goodfellow2014generative}
Goodfellow, I., Pouget-Abadie, J., Mirza, M., Xu, B., Warde-Farley, D., Ozair,
  S., Courville, A., and Bengio, Y. (2014).
\newblock Generative adversarial nets.
\newblock In {\em Advances in neural information processing systems}, pages
  2672--2680.

\bibitem[Graves, 2013]{graves2013generating}
Graves, A. (2013).
\newblock Generating sequences with recurrent neural networks.
\newblock {\em arXiv preprint arXiv:1308.0850}.

\bibitem[He et~al., 2016]{he2016deep}
He, K., Zhang, X., Ren, S., and Sun, J. (2016).
\newblock Deep residual learning for image recognition.
\newblock In {\em Proceedings of the IEEE conference on computer vision and
  pattern recognition}, pages 770--778.

\bibitem[Ho et~al., 2019]{ho2019flow++}
Ho, J., Chen, X., Srinivas, A., Duan, Y., and Abbeel, P. (2019).
\newblock Flow++: Improving flow-based generative models with variational
  dequantization and architecture design.
\newblock {\em arXiv preprint arXiv:1902.00275}.

\bibitem[Huang et~al., 2017]{huang2017densely}
Huang, G., Liu, Z., Van Der~Maaten, L., and Weinberger, K.~Q. (2017).
\newblock Densely connected convolutional networks.
\newblock In {\em Proceedings of the IEEE conference on computer vision and
  pattern recognition}, pages 4700--4708.

\bibitem[Karras et~al., 2017]{karras2017progressive}
Karras, T., Aila, T., Laine, S., and Lehtinen, J. (2017).
\newblock Progressive growing of gans for improved quality, stability, and
  variation.
\newblock {\em arXiv preprint arXiv:1710.10196}.

\bibitem[Kingma and Ba, 2014]{kingma2014adam}
Kingma, D.~P. and Ba, J. (2014).
\newblock Adam: A method for stochastic optimization.
\newblock {\em arXiv preprint arXiv:1412.6980}.

\bibitem[Kingma and Dhariwal, 2018]{kingma2018glow}
Kingma, D.~P. and Dhariwal, P. (2018).
\newblock Glow: Generative flow with invertible 1x1 convolutions.
\newblock In {\em Advances in Neural Information Processing Systems}, pages
  10236--10245.

\bibitem[Kingma et~al., 2016]{kingma2016improved}
Kingma, D.~P., Salimans, T., Jozefowicz, R., Chen, X., Sutskever, I., and
  Welling, M. (2016).
\newblock Improved variational inference with inverse autoregressive flow.
\newblock In {\em Advances in neural information processing systems}, pages
  4743--4751.

\bibitem[Kingma and Welling, 2013]{kingma2013auto}
Kingma, D.~P. and Welling, M. (2013).
\newblock Auto-encoding variational bayes.
\newblock {\em arXiv preprint arXiv:1312.6114}.

\bibitem[Krizhevsky and Hinton, 2009]{krizhevsky2009learning}
Krizhevsky, A. and Hinton, G. (2009).
\newblock Learning multiple layers of features from tiny images.
\newblock Technical report, Citeseer.

\bibitem[Krizhevsky et~al., 2012]{krizhevsky2012imagenet}
Krizhevsky, A., Sutskever, I., and Hinton, G.~E. (2012).
\newblock Imagenet classification with deep convolutional neural networks.
\newblock In {\em Advances in neural information processing systems}, pages
  1097--1105.

\bibitem[LeCun et~al., 1998]{lecun1998gradient}
LeCun, Y., Bottou, L., Bengio, Y., Haffner, P., et~al. (1998).
\newblock Gradient-based learning applied to document recognition.
\newblock {\em Proceedings of the IEEE}, 86(11):2278--2324.

\bibitem[Menick and Kalchbrenner, 2018]{menick2018generating}
Menick, J. and Kalchbrenner, N. (2018).
\newblock Generating high fidelity images with subscale pixel networks and
  multidimensional upscaling.
\newblock {\em arXiv preprint arXiv:1812.01608}.

\bibitem[Oord et~al., 2016a]{oord2016wavenet}
Oord, A. v.~d., Dieleman, S., Zen, H., Simonyan, K., Vinyals, O., Graves, A.,
  Kalchbrenner, N., Senior, A., and Kavukcuoglu, K. (2016a).
\newblock Wavenet: A generative model for raw audio.
\newblock {\em arXiv preprint arXiv:1609.03499}.

\bibitem[Oord et~al., 2016b]{oord2016pixel}
Oord, A. v.~d., Kalchbrenner, N., and Kavukcuoglu, K. (2016b).
\newblock Pixel recurrent neural networks.
\newblock {\em arXiv preprint arXiv:1601.06759}.

\bibitem[Oord et~al., 2017]{oord2017parallel}
Oord, A. v.~d., Li, Y., Babuschkin, I., Simonyan, K., Vinyals, O., Kavukcuoglu,
  K., Driessche, G. v.~d., Lockhart, E., Cobo, L.~C., Stimberg, F., et~al.
  (2017).
\newblock Parallel wavenet: Fast high-fidelity speech synthesis.
\newblock {\em arXiv preprint arXiv:1711.10433}.

\bibitem[Papamakarios et~al., 2017]{papamakarios2017masked}
Papamakarios, G., Pavlakou, T., and Murray, I. (2017).
\newblock Masked autoregressive flow for density estimation.
\newblock In {\em Advances in Neural Information Processing Systems}, pages
  2338--2347.

\bibitem[Parmar et~al., 2018]{parmar2018image}
Parmar, N., Vaswani, A., Uszkoreit, J., Kaiser, {\L}., Shazeer, N., Ku, A., and
  Tran, D. (2018).
\newblock Image transformer.
\newblock {\em arXiv preprint arXiv:1802.05751}.

\bibitem[Radford et~al., 2019]{radford2019language}
Radford, A., Wu, J., Child, R., Luan, D., Amodei, D., and Sutskever, I. (2019).
\newblock Language models are unsupervised multitask learners.

\bibitem[Reed et~al., 2017]{reed2017parallel}
Reed, S., van~den Oord, A., Kalchbrenner, N., Colmenarejo, S.~G., Wang, Z.,
  Chen, Y., Belov, D., and de~Freitas, N. (2017).
\newblock Parallel multiscale autoregressive density estimation.
\newblock In {\em Proceedings of the 34th International Conference on Machine
  Learning-Volume 70}, pages 2912--2921. JMLR. org.

\bibitem[Russakovsky et~al., 2015]{russakovsky2015imagenet}
Russakovsky, O., Deng, J., Su, H., Krause, J., Satheesh, S., Ma, S., Huang, Z.,
  Karpathy, A., Khosla, A., Bernstein, M., et~al. (2015).
\newblock Imagenet large scale visual recognition challenge.
\newblock {\em International journal of computer vision}, 115(3):211--252.

\bibitem[Salimans et~al., 2017]{salimans2017pixelcnn++}
Salimans, T., Karpathy, A., Chen, X., and Kingma, D.~P. (2017).
\newblock Pixelcnn++: Improving the pixelcnn with discretized logistic mixture
  likelihood and other modifications.
\newblock {\em arXiv preprint arXiv:1701.05517}.

\bibitem[van~den Oord et~al., 2016]{van2016conditional}
van~den Oord, A., Kalchbrenner, N., Espeholt, L., Vinyals, O., Graves, A.,
  et~al. (2016).
\newblock Conditional image generation with pixelcnn decoders.
\newblock In {\em Advances in Neural Information Processing Systems}, pages
  4790--4798.

\bibitem[Vaswani et~al., 2017]{vaswani2017attention}
Vaswani, A., Shazeer, N., Parmar, N., Uszkoreit, J., Jones, L., Gomez, A.~N.,
  Kaiser, {\L}., and Polosukhin, I. (2017).
\newblock Attention is all you need.
\newblock In {\em Advances in Neural Information Processing Systems}, pages
  5998--6008.

\end{thebibliography}

\appendix
\section{Optimization details}
\label{appendix:optim}

We use the Adam optimizer~\citep{kingma2014adam} with $\alpha=0.005$ and default $\beta_1$ and $\beta_2$.
Batch size is 256 for MNIST, 32 for all experiemnts on CIFAR-10 and ImageNet 32$\times$32,
24 for ImageNet 64$\times$64, and 8 for CelebA HQ 256$\times$256.
In practice, the weights of invertible $1\times 1$ convolution are possible to become
non-invertible thus interrupts the training (especially on CelebA dataset).
We found it is caused by the increased and then exploded weights of $1\times 1$ convolution during
training. Therefore, for CelebA dataset, we use L2 regularization
for the weights of invertible $1\times 1$ convolution, with $\beta=1.5\times 10^{-8}$. During sampling,
we use the method proposed in~\cite{parmar2018image} to reduce temperature, which often
results in higher-quality images.

\section{Extra Samples}

\begin{figure}[ht]
  \centering
  \includegraphics[width=0.45\linewidth]{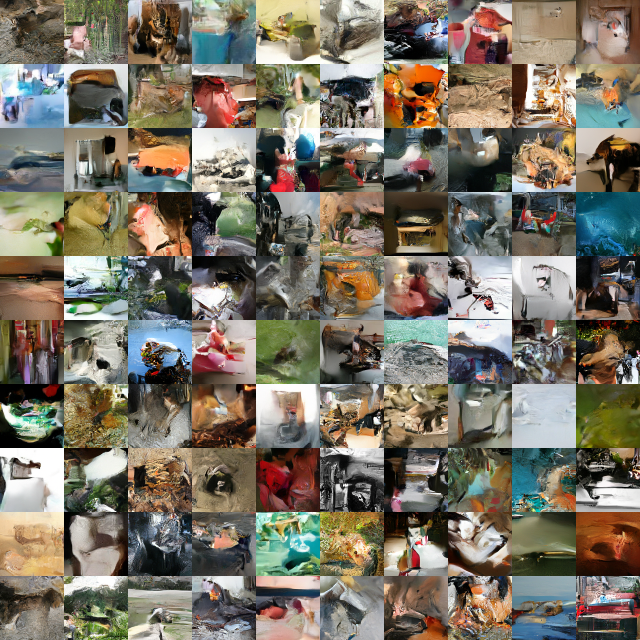}
  \hspace{2.5mm}
  \includegraphics[width=0.45\linewidth]{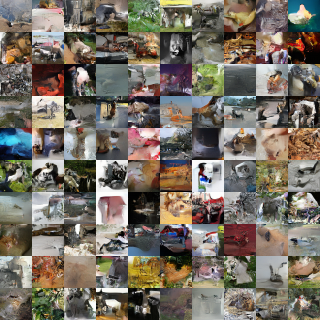}
  \caption{ImageNet 64$\times$64 (left) and 32$\times$32 (right) samples with temperature 1.0}
  \label{fig:imgnet64_extra}
\end{figure}

\begin{figure}[ht]
  \centering
  \includegraphics[width=0.45\linewidth]{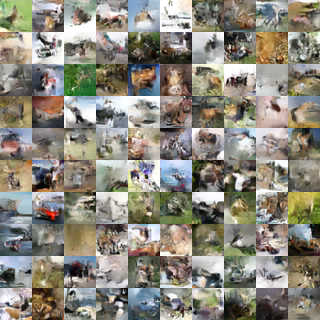}
  \hspace{2.5mm}
  \includegraphics[width=0.45\linewidth]{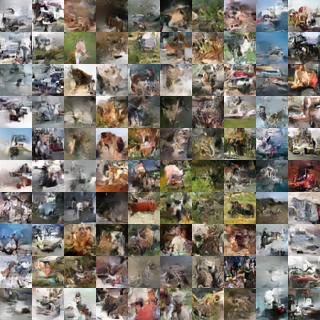}
  \caption{Unconditional (left) and class-conditional (right) samples from CIFAR10 with temperature 1.0}
  \label{fig:cifar10}
\end{figure}

We present extra samples in Fig.~\ref{fig:imgnet64_extra} to~\ref{fig:cifar10}.

\end{document}